# Enriching very large ontologies using the WWW

**Eneko Agirre**[1], **Olatz Ansa**[1], **Eduard Hovy**[2] and **David Martínez**[1]

**Abstract.** This paper explores the possibility to exploit text on the world wide web in order to enrich the concepts in existing ontologies. First, a method to retrieve documents from the WWW related to a concept is described. These document collections are used 1) to construct topic signatures (lists of topically related words) for each concept in WordNet, and 2) to build hierarchical clusters of the concepts (the word senses) that lexicalize a given word. The overall goal is to overcome two shortcomings of WordNet: the lack of topical links among concepts, and the proliferation of senses. Topic signatures are validated on a word sense disambiguation task with good results, which are improved when the hierarchical clusters are used.

## 1 INTRODUCTION

Knowledge acquisition is a long-standing problem in both Artificial Intelligence and Computational Linguistics. Semantic and world knowledge acquisition pose a problem with no simple answer. Huge efforts and investments have been made to build repositories with such knowledge (which we shall call ontologies for simplicity) but with unclear results, e.g. CYC [1], EDR [2], WordNet [3]. WordNet, for instance, has been criticized for its lack of relations between topically related concepts, and the proliferation of word senses.

As an alternative to entirely hand-made repositories, automatic or semi-automatic means have been proposed for the last 30 years. On the one hand, shallow techniques are used to enrich existing ontologies [4] or to induce hierarchies [5], usually analyzing large corpora of texts. On the other hand, deep natural language processing is called for to acquire knowledge from more specialized texts (dictionaries, encyclopedias or domain specific texts) [6][7]. These research lines are complementary; deep understanding would provide specific relations among concepts, whereas shallow techniques could provide generic knowledge about the concepts.

This paper explores the possibility to exploit text on the world wide web in order to enrich WordNet. The first step consists on linking each concept in WordNet to relevant document collections in the web, which are further processed to overcome some of WordNet's shortcomings.

On the one hand, concepts are linked to topically related words. Topically related words form the *topic signature* for each concept in the hierarchy. As in [8][9] we define a topic signature as a family of related terms $\{t, <(w_1,s_1)\ldots(w_i,s_i)\ldots>\}$, where $t$ is the topic (i.e. the target concept) and each $w_i$ is a word associated with the topic, with strength $s_i$. Topic signatures resemble relevancy signatures [10], but are not sentence-based, do not require parsing to construct, and are not suitable for use in information extraction. Topic signatures were originally developed for use in text summarization.

On the other hand, given a word, the concepts that lexicalize it (its word senses) are hierarchically clustered [11], thus tackling sense proliferation in WordNet.

Evaluation of automatically acquired semantic and world knowledge information is not an easy task. In this case we chose to perform task-oriented evaluation, via word sense disambiguation. That is, we used the topic signatures and hierarchical clusters to tag a given occurrence of a word with the intended concept. The benchmark corpus for evaluation is SemCor [12]. Our aim is not to compete with other word sense disambiguation algorithms, but to test whether the acquired knowledge is valid.

This paper describes preliminary experiments. Several aspects could be improved and optimized but we chose to pursue the entire process first, in order to decide whether this approach is feasible and interesting. The resulting topical signatures and hierarchical clusters and their use on word sense disambiguation provide exciting perspectives.

The structure of the paper follows the same spirit: we first explain our method and experiments, and later review some alternatives, shortcomings and improvements. Section two reviews the ontology used and the benchmark corpus for word sense disambiguation. Next the method to build the topic signatures is presented, and a separate section shows the results on a word sense disambiguation task. The clustering method is presented alongside the associated word sense disambiguation results. Related work is discussed in the following section, and finally some conclusions are drawn and further work is outlined.

## 2 BRIEF INTRODUCTION TO WORDNET AND SEMCOR

WordNet is an online lexicon based on psycholinguistic theories [3]. It comprises nouns, verbs, adjectives and adverbs, organized in terms of their meanings around lexical-semantic relations, which include among others, synonymy and antonymy, hypernymy and hyponymy (similar to *is-a* links), meronymy and holonymy (similar to *part-of* links). Lexicalized concepts, represented as sets of synonyms called synsets, are the basic elements of WordNet. The version used in this work, WordNet 1.6, contains 121,962 words and 99,642 concepts.

The noun *boy*, for instance, has 4 word senses, i.e. lexicalized concepts. The set of synonyms for each sense and the gloss is shown below:

*1: male child, boy, child — a youthful male person*
*2: boy — a friendly informal reference to a grown man*
*3: son, boy — a male human offspring*

---
[1] IxA NLP group. University of the Basque Country. 649 pk. 20.080 Donostia. Spain. Email: eneko@si.ehu.es, jipanoso@si.ehu.es, jibmaird@si.ehu.es
[2] USC Information Sciences Institute, 4676 Admiralty Way, Marina del Rey, CA 90292-6695, USA. Email: hovy@isi.edu.

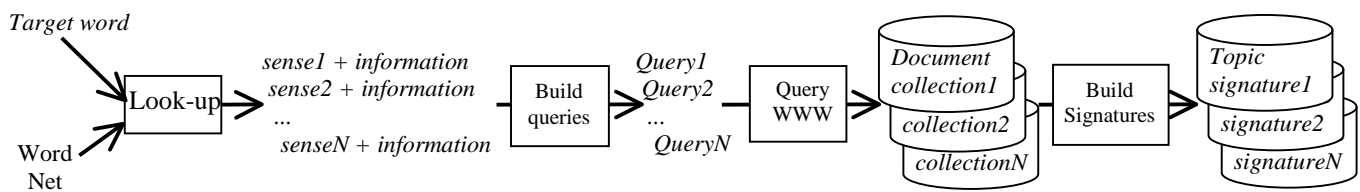

**Figure 1.** Overall design.

*4: boy — offensive term for Black man*

Being one of the most commonly used semantic resources in natural language processing, some of its shortcomings are broadly acknowledged:

1. It lacks explicit links among semantic variant concepts with different part of speech; for instance *paint-to paint* or *song-to sing* are not related.

2. Topically related concepts are not explicitly related: there is no link between pairs like *bat–baseball, fork–dinner, farm–chicken,* etc.

3. The proliferation of word sense distinctions in WordNet, which is difficult to justify and use in practical terms, since many of the distinctions are unclear. *Line* for instance has 32 word senses. This makes it very difficult to perform automatic word sense disambiguation.

This paper shows how to build lists of words that are topically related to a topic (a concept). These lists can be used to overcome the shortcomings just mentioned. In particular we show how to address the third issue, using the lists of words to cluster word senses according to the topic.

SemCor [12] is a corpus in which word sense tags (which correspond to WordNet concepts) have been manually included for all open-class words in a 360,000-word subset of the Brown Corpus. We use SemCor to evaluate the topic signatures in a word sense disambiguation task. In order to choose a few nouns to perform our experiments, we focused on a random set of 20 nouns which occur at least 100 times in SemCor. The set comprises commonly used nouns like *boy, child, action, accident, church*, etc. These nouns are highly polysemous, with 6.3 senses on average.

## 3 BUILDING TOPIC SIGNATURES FOR THE CONCEPTS IN WORDNET

In this work we want to collect for each concept in WordNet the words that appear most distinctively in texts related to it. That is, we aim at constructing lists of closely related words for each concept. For example, WordNet provides two possible word senses or concepts for the noun *waiter*:

*1: waiter, server — a person whose occupation is to serve at table (as in a restaurant)*
*2: waiter — a person who waits or awaits*

For each of these concepts we would expect to obtain two lists with words like the following:

*1: restaurant, menu, waitress, dinner, lunch, counter,* etc.
*2: hospital, station, airport, boyfriend, girlfriend, cigarette,* etc.

The strategy to build such lists is the following (cf. Figure 1). We first exploit the information in WordNet to build queries, which are used to search in the Internet those texts related to the given word sense. We organize the texts in collections, one collection per word sense. For each collection we extract the words and their frequencies, and compare them with the data in the other collections. The words that have a distinctive frequency for one of the collections are collected in a list, which constitutes the topic signature for each word sense.

The steps are further explained below.

### 3.1 Building the queries

The original goal is to retrieve from the web all documents related to an ontology concept. If we assume that such documents have to contain the words that lexicalize the concept, the task can be reduced to classifying all documents where a given word occurs into a number of collections of documents, one collection per word sense. If a document cannot be classified, it would be assigned to an additional collection.

The goal as phrased above is unattainable, because of the huge amount of documents involved. Most of words get millions of hits: *boy* would involve retrieving 2,325,355 documents, *church* 6,243,775, etc. Perhaps in the future a more ambitious approach could be tried, but at present we cannot aim at classifying those enormous collections. Instead, we construct queries, one per concept, which are fed to a search engine. Each query will retrieve the documents related to that concept.

The queries are constructed using the information in the ontology. In the case of WordNet each concept can include the following data: words that lexicalize the concept (synonyms), a gloss and examples, hypernyms, hyponyms, meronyms, holonyms and attributes. Altogether a wealth of related words is available, which we shall call cuewords. If a document contains a high number of such cuewords around the target word, we can conclude that the target word corresponds to the target concept. The cuewords are used to build a query which is fed into a search engine, retrieving the collection of related documents.

As we try to constrain the retrieved documents to the 'purest' documents, we build the queries for each word sense trying to discard documents that could belong to more than one sense. For instance, the query for word $x$ in word sense i (being $j,k$ other word senses for $x$) is constructed as follows:

($x$ AND ($cueword_{1,i}$ OR $cueword_{2,i}$ ...)
    AND NOT ($cueword_{1,j}$ OR $cueword_{2,j}$ ... OR
        $cueword_{1,k}$ OR $cueword_{2,k}$ ...)

where $cueword_{l,m}$ stands for the cueword $l$ of word sense $m$. This boolean query searches for documents that contain the target word together with one of the cuewords of the target concept, but do not contain any of the cuewords of the remaining concepts. If a cueword appears in the information relative to more than one sense, it is discarded.

Deciding which of the cuewords to use, and when, is not an easy task. For instance, nouns in the definition are preferable to the other parts of speech, monosemous cuewords are more valuable than polysemous ones, synonyms provide stronger evidence than meronyms, other concepts in the hierarchy can also be used, etc. After some preliminary tests, we decided to experiment with all available information: synonyms, hypernyms, hyponyms, coordinate sisters, meronyms, holonyms and nouns in the definition. Table 1 shows part of the information available for sense 1 of boy.

**Table 1.** Information for sense 1 of boy.

| | |
|---|---|
| synonyms | *male child, child* |
| gloss | *a youthful male person* |
| hypernyms | *male, male person* |
| hyponyms | *altar boy, ball boy, bat boy, cub, lad, laddie, sonny, sonny boy, boy scout, farm boy, plowboy, ...* |
| coordinate systers | *chap, fellow, lad, gent, fella, blighter, cuss, foster brother, male child, boy, child, man, adult male, ...* |

The query for sense 1 of boy would include the above cuewords plus the negation for the cuewords of the other senses. An excerpt of the query:

*(boy AND ('altar boy' OR 'ball boy' OR ...OR 'male person)*
    *AND NOT ('man'... OR 'broth of a boy' OR  # sense 2*
                *'son' OR... OR 'mama's boy' OR  # sense 3*
                *'nigger' OR ... OR 'black')  # sense 4*

## 3.2 Search the internet

Once the queries are constructed we can use a number of different search engines. We started to use just the first 100 documents from a list of search engines. This could bias the documents, and some could be retrieved repeatedly. Therefore, unlike [13], we decided to use only one search engine, the most comprehensive search engine at the time, AltaVista [14]. AltaVista allows complex queries which were not possible in some of the other web search engines.

The number of documents retrieved for the 20 words amounts to the tens of thousands, taking more that one gigabyte of disk space once compressed, and 9 days of constant internet access. For instance, it took 3 hours and a half to retrieve the 1,217 documents for the four senses of *boy*, which took 100 megabytes once compressed.

## 3.3 Build topic signatures

The document collections retrieved in step 3.2 are used to build the topic signatures. The documents are processed in order to extract the words in the text. We did not perform any normalization; the words are collected as they stand. The words are counted and a vector is formed with all words and their frequencies in the document collection. We thus obtain one vector for each collection, that is, one vector for each word sense of the target word.

In order to measure which words appear distinctively in one collection in respect to the others, a signature function was selected based on previous experiments [15][13]. We needed a function that would give high values for terms that appear more frequently than expected in a given collection. The signature function that we used is $\chi^2$, which we will define next.

The vector $vf_i$ contains all the words and their frequencies in the document collection $i$, and is constituted by pairs $(word_j, freq_{i,j})$, that is, one word $j$ and the frequency of the word $j$ in the document collection $i$. We want to construct another vector $vx_i$ with pairs $(word_j, w_{i,j})$ where $w_{i,j}$ is the $\chi^2$ value for the word $j$ in the document collection $i$ (cf. Equation 1).

$$w_{i,j} = \begin{cases} \dfrac{(freq_{i,j} - m_{i,j})}{m_{i,j}} & \text{if } freq_{i,j} > m_{i,j} \\ 0 & \text{otherwise} \end{cases} \quad (1)$$

**Table 2.** Top words in signatures for three senses of boy.

| Boy1 | Boy2 | Boy3 |
|---|---|---|
| (child 9854) | (gay 7474) | (human 5023) |
| (Child 5979) | (reference 5154) | (son 4898) |
| (person 4671) | (tpd-results 3930) | (Human 3055) |
| (anything.com 3702) | (sec 3917) | (Soup 1852) |
| (Opportunities 1808) | (gay 2906) | (interactive 1842) |
| (Insurance 1796) | (Xena 1604) | (hyperinstrument 1841) |
| (children 1458) | (male 1370) | (Son 1564) |
| (Girl 1236) | (ADD 1304) | (clips 1007) |
| (Person 1093) | (storing 1297) | (father 918) |
| (Careguide 918) | (photos 1203) | (man-child 689) |
| (Spend 839) | (merr 1077) | (measure 681 ) |
| (Wash 821) | (accept 1071) | (focus 555) |
| (enriching 774) | (PNorsen 1056) | (research 532) |
| (prizes 708) | (software 1021) | (show 461) |
| (Scouts 683) | (adult 983) | (Teller 456) |
| (Guides 631) | (penny 943) | (Yo-Yo 455) |
| (Helps 614) | (PAGE 849) | (modalities 450) |
| (Christmas 525) | (Sex 835) | (performers 450) |
| (male 523) | (Internet 725) | (senses 450) |
| (address 504) | (studs 692) | (magicians 448) |
| (paid 472) | (porno 675) | (percussion 439) |
| (age 470) | (naked 616) | (mother 437) |
| (mother 468) | (erotic 611) | (entertainment 391) |
| ...up to 6.4 Mbytes | ...up to 4.4 Mbytes | ... up to 4.7 Mbytes |

Equation 2 defines $m_{i,j}$, the expected mean of word $j$ in document $i$.

$$m_{i,j} = \frac{\Sigma_j freq_{i,j} \; \Sigma_j freq_{i,j}}{\Sigma_{i,j} freq_{i,j}} \quad (2)$$

When computing the $\chi^2$ values, the frequencies in the target document collection are compared with the rest of the document collection, which we call the *contrast set*. In this case the contrast set is formed by the other word senses. Excerpts from the signatures for boy are shown in Table 2.

## 4 APPLY SIGNATURES FOR WORD SENSE DISAMBIGUATION

The goal of this experiment is to evaluate the automatically constructed topic signatures, not to compete against other word sense disambiguation algorithms. If topic signatures yield good results in word sense disambiguation, it would mean that topic signatures have correct information, and that they are useful for word sense disambiguation. Given the following sentence from SemCor, a word sense disambiguation algorithm should decide that the intended meaning for *waiter* is that of a restaurant employee:

*"There was a brief interruption while one of O'Banion's men jerked out both his guns and threatened to shoot a waiter who was pestering him for a tip."*

Word sense disambiguation is a very active research area (cf. [16] for a good review of the state of the art). Present word sense disambiguation systems use a variety of information sources [17] which play an important role, such us collocations, selectional restrictions, topic and domain information, co-occurrence relations, etc. Topic signatures constitute one source of evidence, but do not replace the others. Therefore, we do not expect impressive results.

The word sense disambiguation algorithm is straightforward. Given an occurrence of the target word in the text we collect the words in its context, and for each word sense we retrieve the $\chi^2$ values for the context words in the corresponding topic signature.

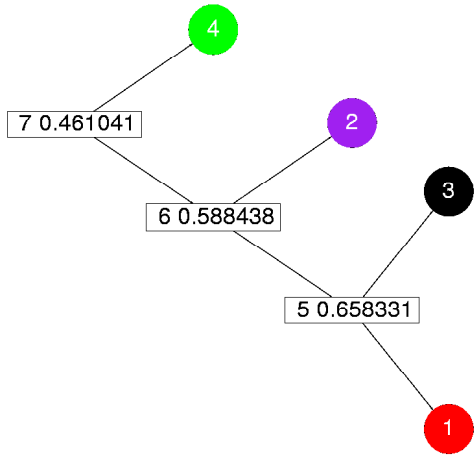

**Figure 2:** Hierarchy for the word senses of boy

For each word sense we add these $\chi^2$ values, and then select the word sense with the highest value. Different context sizes have been tested in the literature, and large windows have proved to be useful for topical word sense disambiguation [18]. We chose a window of 100 words around the target.

In order to compare our results, we computed a number of baselines. First of all choosing the sense at random (*ran*). We also constructed lists of related words using WordNet, in order to compare their performance with that of the signatures: the list of synonyms (*Syn*), these plus the content words in the definitions (*S+def*), and these plus the hyponyms, hypernyms and meronyms (*S+all*). The algorithm to use these lists is the same as for the topic signatures.

Table 3 shows the results for the selected nouns. The number of senses attested in SemCor[3] (*#s*) and the number of occurrences of the word in SemCor (*#occ*) are also presented. The results are given as precision, that is, the number of successful tags divided by the total number of occurrences. A precision of one would mean that all occurrences of the word are correctly tagged.

The results show that, the precision of the signature-based word sense disambiguation (*Sign* column) is well above the precision for random selection (a few exceptions are in bold), and, that, overall, it outperforms the other WordNet-based lists of words (the winner for each word is in bold). This proves that topic signatures managed to learn topic information that was not originally present in WordNet. This information is overly correct, but in some cases introduces noise and the performance degrades even below the random baseline (e.g. *action, hour*).

## 5 CLUSTERING WORD SENSES

In principle we could try to cluster all the concepts in WordNet, comparing their topic signatures, but instead we experimented with clustering just the concepts that belong to a given word (its word senses). As we mentioned in Section 2, WordNet makes very fine distinctions between word senses, and suffers excessive word sense proliferation.

For many practical applications we can ignore some of the sense distinctions. For instance, all of the senses for boy are persons.

---

[3] Some word senses never occur in SemCor. We did not take those senses into account.

**Table 3.** Word sense disambiguation results.

| Word | #s | #occ | Ran | Syn | S+def | S+all | Sign |
|---|---|---|---|---|---|---|---|
| Accident | 2 | 12 | 0.50 | 0.00 | **0.56** | **0.71** | 0.50 |
| Action | 8 | 130 | **0.12** | 0.00 | **0.05** | **0.29** | 0.02 |
| Age | 3 | 104 | 0.33 | 0.01 | 0.04 | 0.03 | **0.60** |
| Amount | 4 | 103 | 0.25 | 0.22 | 0.27 | 0.30 | **0.50** |
| Band | 7 | 21 | 0.14 | 0.11 | 0.13 | **0.28** | 0.25 |
| Boy | 4 | 169 | 0.25 | 0.45 | 0.37 | 0.59 | **0.66** |
| Cell | 3 | 116 | 0.33 | 0.00 | 0.37 | 0.36 | **0.59** |
| Child | 2 | 206 | **0.50** | 0.37 | **0.47** | **0.43** | 0.29 |
| Church | 3 | 128 | 0.33 | 0.28 | **0.50** | 0.46 | 0.45 |
| Difference | 5 | 112 | **0.20** | 0.02 | **0.28** | **0.35** | 0.17 |
| Door | 4 | 138 | **0.25** | 0.05 | **0.24** | **0.26** | 0.04 |
| Experience | 3 | 125 | 0.33 | 0.22 | **0.42** | 0.35 | **0.42** |
| Fact | 4 | 124 | 0.25 | 0.02 | 0.48 | 0.58 | **0.82** |
| Family | 6 | 135 | 0.17 | 0.12 | 0.18 | 0.15 | **0.36** |
| Girl | 5 | 152 | 0.20 | **0.34** | 0.21 | **0.33** | 0.25 |
| History | 5 | 104 | **0.20** | 0.06 | 0.16 | 0.17 | 0.18 |
| Hour | 2 | 110 | **0.50** | 0.21 | **0.63** | 0.38 | 0.40 |
| Information | 3 | 146 | 0.33 | 0.00 | 0.12 | 0.64 | **0.66** |
| Plant | 2 | 99 | 0.50 | 0.30 | 0.42 | 0.45 | **0.82** |
| World | 8 | 210 | 0.12 | 0.09 | 0.18 | 0.19 | **0.34** |
| **Overall** | 83 | 2444 | 0.28 | 0.16 | 0.30 | 0.36 | **0.41** |

Two of the senses refer to young boys while two of them refer to grown males. '*Boy as a young person*' would tend to appear in a certain kind of documents, while '*boy as a grown man*' in others, and '*boy as a colored person*' in yet other documents.

In this work, as in [15][13], we tried to compare the overlap between the signatures by simply counting shared words, but this did not yield interesting results. Instead we used binary hierarchical clustering directly on the retrieved documents [11]. We experimented with various distance metrics and clustering methods but the results did not vary substantially: slink [19], clink [20], median, and Ward's method [21]. Some of the resulting hierarchies were analyzed by hand and they were coherent according to our own intuitions. For instance Figure 2 shows that the *young* and *offspring* senses of boy (nodes 1 and 3) are the closest (similarity of 0.65), while the *informal* (node 2) and *colored* (node 4) senses are further apart. The contexts for the *colored* sense are the least similar to the others (0.46).

### 5.1 Evaluation of word sense clusters on a word sense disambiguation task

Hand evaluation of the hierarchies is a difficult task, and very hard to define [11]. As before, we preferred to evaluate them on a word sense disambiguation task. We devised two methods to apply the hierarchies and topic signatures to word sense disambiguation:

1. Use the original topic signatures. In each branch of the hierarchy we combine all the signatures for the word senses in the branch, and choose the highest ranking branch. For instance, when disambiguating *boy*, we first choose between *boy4* and the rest: *boy1, boy2, boy3* (cf. Figure 2). Given a occurrence, the evidence for b*oy1, boy2, boy3* is combined, and compared to the evidence for *boy4*. The winning branch is chosen. If *boy4* is discarded, then the combined evidence for *boy1, boy3* is compared to that of *boy2*. If *boy2* gets more evidence, that is the chosen sense.

2. Build new topic signatures for the existing clusters. The document collections for all the word senses in the branch are merged and new $\chi^2$ values are computed for each cluster in the hierarchy. For instance, at the first level we would have a topic signature for *boy4* and another for the merged collections of *boy1*, *boy2* and *boy3*. At the second level we would have a topic signature for *boy2* and another for *boy1*, *boy3*.

The word sense disambiguation algorithm can be applied at different levels of granularity, similar to decision trees. At the first level it chooses to differentiate between *boy4* and the rest, at the second level among *boy4*, *boy2* and *boy1-3*, and at the third level it disambiguates the finest-grained senses.

Instead of evaluating the set of all nouns, we focused on three nouns: *boy*, *cell* and *church*. The results are shown in Table 4. The second column shows the number of senses. The signature results for the original sense distinctions (cf. Table 3) are shown in the second column. The results for the signature and hierarchy combination are shown according to the sense-distinctions: the fine column shows the results using the hierarchy for the finest sense distinctions, the medium column corresponds to the medium sized clusters, and the coarse level corresponds to the coarsest clusters, i.e., all senses clustered in two groups. For each level, three results are given: the random baseline, the results using the original topic signatures and the hierarchy, and the results with the new topic signatures computed over the clusters (best results for in bold).

**Table 4:** Results using hierarchies and word sense clusters

| Word | # | Sign Orig | Signature & Hierarchy | | | | | | | |
|---|---|---|---|---|---|---|---|---|---|---|
| | | | Fine | | | Medium | | | Coarse | | |
| | | | Rand | Orig | New | Rand | Orig | New | Rand | Orig | New |
| Boy | 4 | 0.66 | 0.25 | **0.68** | 0.38 | 0.33 | **0.83** | 0.67 | 0.50 | **0.99** | 0.99 |
| Cell | 3 | 0.59 | 0.33 | **0.62** | 0.52 | - | - | - | 0.50 | 0.52 | **0.96** |
| Church | 3 | 0.45 | 0.33 | 0.48 | **0.54** | - | - | - | 0.50 | 0.77 | **0.90** |

The results show that the information contained in the hierarchy helps improve the precision obtained without hierarchies, even at the fine level. For coarser sense distinctions it exceeds 0.90 precision. Regarding the way to apply the hierarchy, the results are not conclusive. Further experiments would be needed to show whether it is useful or not to compute new topic signatures for each cluster.

# 6 DISCUSSION AND COMPARISON WITH RELATED WORK

The work here presented involves different areas of research. We will focus on the method to build topic signatures, the method to cluster the concepts and how the document collection for each word sense is constructed.

## 6.1 Building topic signatures

Topic signatures were an extension of relevancy signatures [10] developed for text summarization [15]. To identify topics in documents, [15] constructed topic signatures from 16,137 documents classified into 32 topics of interest. His topic signature construction method is similar to ours, except that he used *tf.idf* for term weighting. In subsequent work, Hovy and Junk [13] explored several alternative weighting schemes in a topic identification task, finding that $\chi^2$ provided better results than *tf.idf* or *tf*, and that specific combinations of $\chi^2$ and Latent Semantic Analysis provided even better results on clean training data. Lin and Hovy [9] use a likelihood ratio from maximum likelihood estimates that achieves even better performance on clean data. However, their experiments with text extracted from the web proved somewhat disappointing, like the ones reported here.

In general, documents retrieved from the web introduce a certain amount of noise into signatures. The results are still useful to identify the word sense of the target words, as our results show, but a hand evaluation of them is rather worrying. We concluded that the cause of the poor quality does not come from the procedure to build the signatures, but rather from the quality of the documents retrieved (c.f. Section 6.3).

## 6.2 Concept Clustering

Traditional clustering techniques [11] are difficult to apply to concepts in ontologies. The reason is that the usual clustering methods are based on statistical word co-occurrence data, and not on concept co-occurrence data (which is not available at present). The method presented in this paper uses the fact that concepts are linked to document collections. Usual document clustering techniques are applied to document collections, effectively clustering the associated concepts. This clustering method tackles the word sense proliferation WordNet.

The evaluation and validation of the word sense clusters is difficult [11]. We chose to evaluate the performance of the clusters in a word sense disambiguation task, showing that the clusters are useful to improve the results; enhanced precision for the fine-level sense distinctions, and over 90% precision for the coarse level.

## 6.3 Searching the internet for concepts

The core component of the method explored in this paper is the technique to link documents in the web to concepts in an ontology. Recently, some methods have been explored to automatically retrieve examples for concepts from large corpora and the internet. Leacock et al. [22] use a strategy based on the monosemous relatives of WordNet concepts to retrieve examples from a 30 million word corpus. As their goal is to find 100 examples for each word sense of a given word, they prefer close relatives such us synonyms or hyponym collocations that contain the target hyponym. If enough examples are not found, they also use other hyponyms, sisters and hypernyms. The examples were used to train a supervised word sense disambiguation algorithm with very good results, but no provision was made to enrich WordNet with them. The main shortcoming of this strategy is that limiting the search to monosemous relatives, only 65% of the concepts under study could get training examples.

Mihalcea and Mondovan [23] present a similar work which tries to improve the previous method. When a monosemous synonym for a given concept is not found, additional information from the definition of the concept is used, in the form of defining phrases constructed after parsing and processing the definition. The whole internet is used as a corpus, using a search engine to retrieve the examples. Four procedures are defined to query the search engine in order: use monosemous synonyms, use the defining phrases, use synonyms with the AND operator and words from the defining phrase with the NEAR operator, and lastly, use synonyms and words from the defining phrases with the AND operator. The procedures are sorted by preference, and one procedure is only applied if the previous one fails to retrieve any examples. 20 words totaling 120 senses were chosen, and an average of 670 examples could be retrieved for each word sense. The top 10 examples for each word sense were hand-checked and 91% were found correct.

Both these methods focus on obtaining training examples. In contrast, our method aims at getting documents related to the concept. This allows us to be less constraining; the more documents the better, because that allows to found more distinctively co-occurring terms. That is why we chose to use all close relatives for a given concept, in contrast to [22] which only focuses on monosemous relatives, and [23], which uses synonyms and a different strategy to process the gloss. Another difference is that our method forbids the cuewords of the rest of the senses.

We have found that searching the web is the weakest point of our method. The quality and performance of the topic signatures and clusters depends on the quality and number of the retrieved documents, and our query strategy is not entirely satisfactory. On the one hand some kind of balance is needed. For some querying strategies some word senses do not get any document, and with other strategies too many and less relevant documents are retrieved. On the other hand the web is not a balanced corpus (e.g. the sexual content in the topic signatures for boy). Besides, many documents are short indexes or cover pages, with little text on them. In this sense, the query construction has to be improved and some filtering techniques should be devised.

Other important consideration about searching the internet is that technical features have to be taken in consideration. For instance, our system had some timeout parameters, meaning that the retrieval delay of the documents (caused by the hour, workload, localization of server, etc.) could affect the results.

## 7 CONCLUSIONS AND FURTHER RESEARCH

We have introduced an automatic method to enrich very large ontologies, e.g. WordNet, that uses the huge amount of documents in the world wide web. The core of our method is a technique to link document collections from the web to concepts, which allows to alleviate some of the main problems acknowledged in WordNet; lack of relations between topically related concepts, and the proliferation of word senses. We show in practice that the document collections can be used 1) to create topic signatures (lists of words that are topically related to the concept) for each WordNet concept, and, 2) given a word, to cluster the concepts that lexicalize it (its word senses), thus tackling sense proliferation. In order to validate the topic signatures and word sense clusters, we demonstrate that they contain information which is useful in a word sense disambiguation task.

This work combines several techniques, and we chose to pursue the whole method from start to end. This strategy left much room for improvement in all steps. Both signature construction and clustering seem to be satisfactory, as other work has also shown. In particular, nice clean signatures are obtained when constructing topic signatures from topically organized documents. On the contrary, topic signatures extracted from the web seem to be dirtier.

We think that, in this work, the main obstacle to get clean signatures comes from the method to link concepts and relevant documents from the web. The causes are basically two. First, the difficulty to retrieve documents relevant to one and only one concept. The query construction has to be improved and carefully fine-tuned to overcome this problem. Second, the wild and noisy nature of the texts in the web, with its high bias towards some topics, high number of not really textual documents e.g., indexes., etc. Some filtering techniques have to be applied in order to get documents with less bias and more content.

Cleaner topic signatures open the avenue for interesting ontology enhancements, as they provide concepts with rich topical information. For instance, similarity between topic signatures could be used to find out topically related concepts, the clustering strategy could be extended to all concepts rather that just the concepts that lexicalize the same word, etc. Besides, word sense disambiguation methods could profit from these richer ontologies, and improve word sense disambiguation performance.

## ACKNOWLEDGEMENTS

We would like to thank the referees for their fruitful comments. Part of the work was done while Eneko Agirre was visiting ISI, funded by the Basque Government.